\documentclass[conference,letter,twocolumn]{IEEEtran}

\usepackage{subfigure} 
\usepackage{amsmath} 
\usepackage{amssymb}  
\usepackage{graphicx}  
\pdfoutput=1
\newif\ifpdf\ifx\pdfoutput\undefined\pdffalse\else\pdfoutput=1\pdftrue\fi

\begin{document}

\title{\LARGE \bf Multi-Robot Searching Algorithm \\Using Levy Flight and Artificial Potential Field}

\author{\authorblockN{ $^1$Donny K. Sutantyo, $^1$Serge Kernbach, $^2$Valentin A. Nepomnyashchikh and $^1$Paul Levi}
\vspace{1mm}
\authorblockA{$^{1}$Institute of Parallel and Distributed System (IPVS), University of Stuttgart,\\
Universit\"atstrasse 38, Stuttgart, Germany\\
\emph{\{sutantdy,serge.kernbach,paul.levi\}@ipvs.uni-stuttgart.de},\\[1mm]
$^{2}$Institute for Biology of Inland Waters, Russian Academy of Sciences, \\
152 742 Borok, Yaroslavskaja obl., Russia\\
\emph{nepom@ibiw.yaroslavl.ru}\\[-2mm]
}
}

\maketitle

\begin{abstract}
\footnote{Appeared in the eighth IEEE International Workshop on Safety, Security, and Rescue Robotics (SSRR-2010), Bremen, Germany, 26-30 July 2010}An efficient search algorithm is very crucial in robotic area, especially for exploration missions, where the target availability is unknown and the condition of the environment is highly unpredictable. In a very large environment, it is not sufficient to scan an area or volume by a single robot, multiple robots should be involved to perform the collective exploration. In this paper, we propose to combine bio-inspired search algorithm called Levy flight and artificial potential field method to perform an efficient searching algorithm for multi-robot applications. The main focus of this work is not only to prove the concept or to measure the efficiency of the algorithm by experiments, but also to develop an appropriate generic framework to be implemented both in simulation and on real robotic platforms. Several experiments, which compare different search algorithms, are also performed.
\end{abstract}

\vspace{0.2cm} {\bfseries{\textit{\small Keywords}}}: \ {\it{Multi-Robot, Levy Flight, Artificial Potential Field, Random Search Algorithm}}

\IEEEpeerreviewmaketitle

\section{INTRODUCTION}

Unmanned area exploration is crucial for investigating biological species, monitoring pollution, disaster warning system, and search-rescue mission. Autonomous robot equipped with sensing peripherals is deployed in the environment to find the object of interest, i.e., fire spots in the jungle, missing black box from a crashed airplane, or to measure a concentration of hazardous materials. However, in a very large environment, it is not sufficient to scan an area or volume with a single robot. Many autonomous robots, which have wireless communication capabilities are deployed to improve searching accuracy. Therefore, beside having an optimal search strategy, the whole system becomes wireless mobile sensor network that is able to perform distributed sensing in a dynamic environment. If the number of autonomous robots is large, the swarm phenomena can also be achieved.

There are many bio-inspired examples of how to realize efficient foraging. Animal optimize its search for food with physical and biological constraints, which restrict the behavior. Evolutionary process through natural selection led over time to highly efficient and optimal foraging strategies [22], e.g. how lobster localize and track odor plumes or bacterial chemotaxis mechanism used by e-coli to response the nutrition concentration gradient [2]. In case that the forager can only detect the randomly located objects in limited vicinity, random walk is performed to explore the environment. General question related to the best statistical strategy of optimizing the random search has been addressed by many scientist. In [2],[3],[4],[5], it is shown that the random search efficiency depends on the probability distribution of the flight length taken by the forager. When the target sites are sparsely and randomly distributed, the optimum strategy is a specialized random walks movement, called Levy flight [4]. Levy flight is a random walk mechanism that has the Levy probability distribution function in determining the length of the walk. By performing Levy flight, forager optimizes the number of targets encountered versus the traveled distance. The idea is that the probability of returning to the previous site is smaller compared with other random walk mechanism [4]. This Levy flight motion has been found among various organisms, such as marine predators, fruit flies, and honey bee [2].

\begin{figure}[ht]
\centering
\subfigure[]{\includegraphics[width=.18\textwidth]{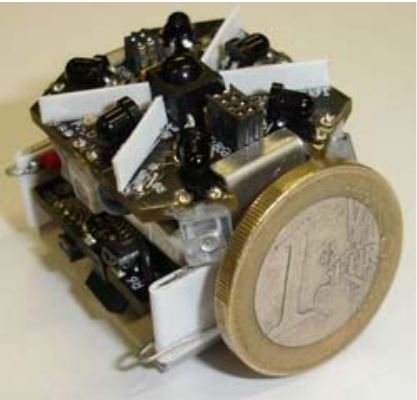}}~
\subfigure[]{\includegraphics[width=.25\textwidth]{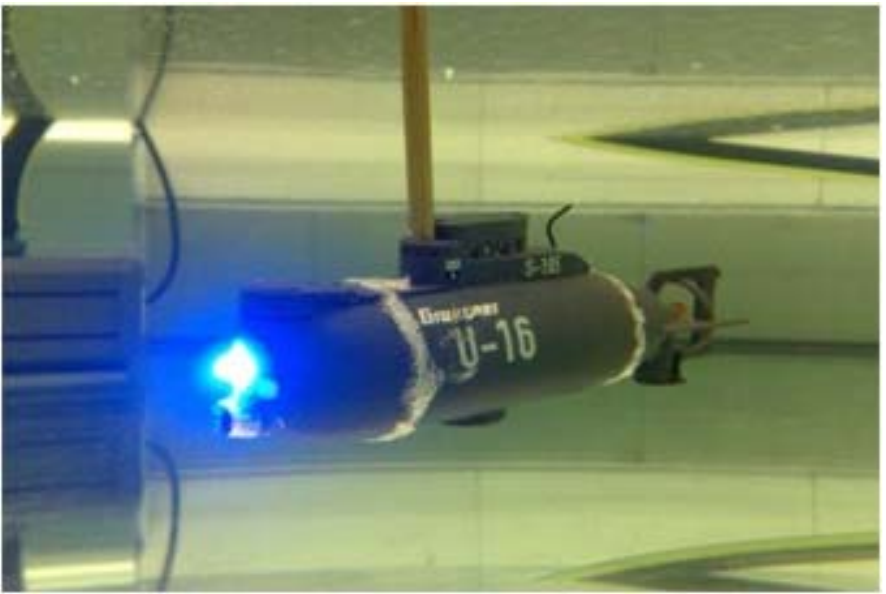}}
\caption{Swarm robot platform. \textbf{(a)} Micro-robot Jasmine IIIp; \textbf{(b)} ANGELS Underwater Test Platform.
\label{fig:jasmine}}
\end{figure}

In this paper, we proposed to combine Levy flight mechanism and artificial potential field method to perform an optimum bio-inspired foraging strategy. The Levy flight algorithm will generate the length of the movement, while the artificial potential field will improve the dispersion of the deployed robot by generating repulsion forces among robots. The intention of finding an optimum random searching algorithm came from ANGELS and SwarmRobot projects [18],[10],[20]. In one part of ANGELS project scenario, several underwater mobile robots will be applied for searching an object of interest in underwater applications. Since the dynamic and kinematic behavior of the mobile robot are complex and difficult to be measured, a simulation platform is used for investigating the feasibility and the efficiency of the random search implementation. The simulation environment also simplify the observation during experiment, since the ANGELS and Jasmine robot, see Fig.~\ref{fig:jasmine}, do not have the global localization system to achieve the online coordinate position during the experiment.
In this work, a simplification of 3D underwater environment from the real ANGELS scenario into 2D simulative surface is applied for the first random search experiments. Once the efficient random search algorithm is developed, then it can be easily implemented in the real Jasmine robot platform for surface application by modifying several parameters, and later can be extended and improved for the real 3D environment with underwater robot. In order to measure the performance of the algorithm, several experiments are provided by changing several parameters in the algorithm. Furthermore, comparing the algorithm with a usual random search method is also necessary.

The main focus of this paper is not only to measure the efficiency of the searching algorithm by using simulation, but also to find an appropriate solution in designing a generic framework to implement the Levy flight algorithm for robotic applications. However, the generic framework must be appropriate both for simulation and for real robots. Furthermore, since the architecture of the swarm robot should be simple and small, an implementation that has only light computational requirements is crucial.

The rest of this paper is structured in the following way. The Secs.~\ref{sec:levy} and \ref{sec:petentF} describe theoretical approaches underlying the Levy flight and artificial potential field. Secs.~\ref{sec:imp} and \ref{sec:exp} are devoted to implementation and experiments, whereas Sec.~\ref{sec:conlusion} concludes this work.

\section{Levy Flight Random Search}
\label{sec:levy}

Biological creatures performs different exploration activities: e.g. to search for sources of food that are not visible in the immediate vicinity of the animal, to search for a new site, to search for a mate, or to avoid predators. The knowledge about the condition of the environment implies the complexity of the searching strategy. If the environment is unchanging and highly predictable, animal is able to develop knowledge of where to forage. If the resource availability is unknown (according to the perception of animals) and the condition of the environment is unpredictable, animals have to conduct non-oriented searches with little or no prior knowledge of where and how resources are distributed. Regarding the physical and biological constraint, the capability to find the resources efficiently will minimize the risk of starvation.

Brownian walk and Levy flight are two well known biological random search. For many years, the Brownian walk was the most used model for describing non-oriented animal movement [1].The main difference between Brownian random walk and Levy flight is that Levy flight uses Levy probability distribution that has power tail instead of using normal Gaussian distribution, that makes the probability of returning to previously visited site is smaller, and therefore advantageous when target sites are sparsely and randomly distributed [1], [3]. The other difference came from the intention in deciding the length of the movements. In the Brownian walk, the scale of the movement is determined by the organism. However, in the Levy flight, the larger scale of the movement is defined by the distribution of the object of interest, thus makes Levy flight more flexible and adaptive to the environmental changes [3].
\begin{figure}[ht]
\centering
\includegraphics[width=.4\textwidth]{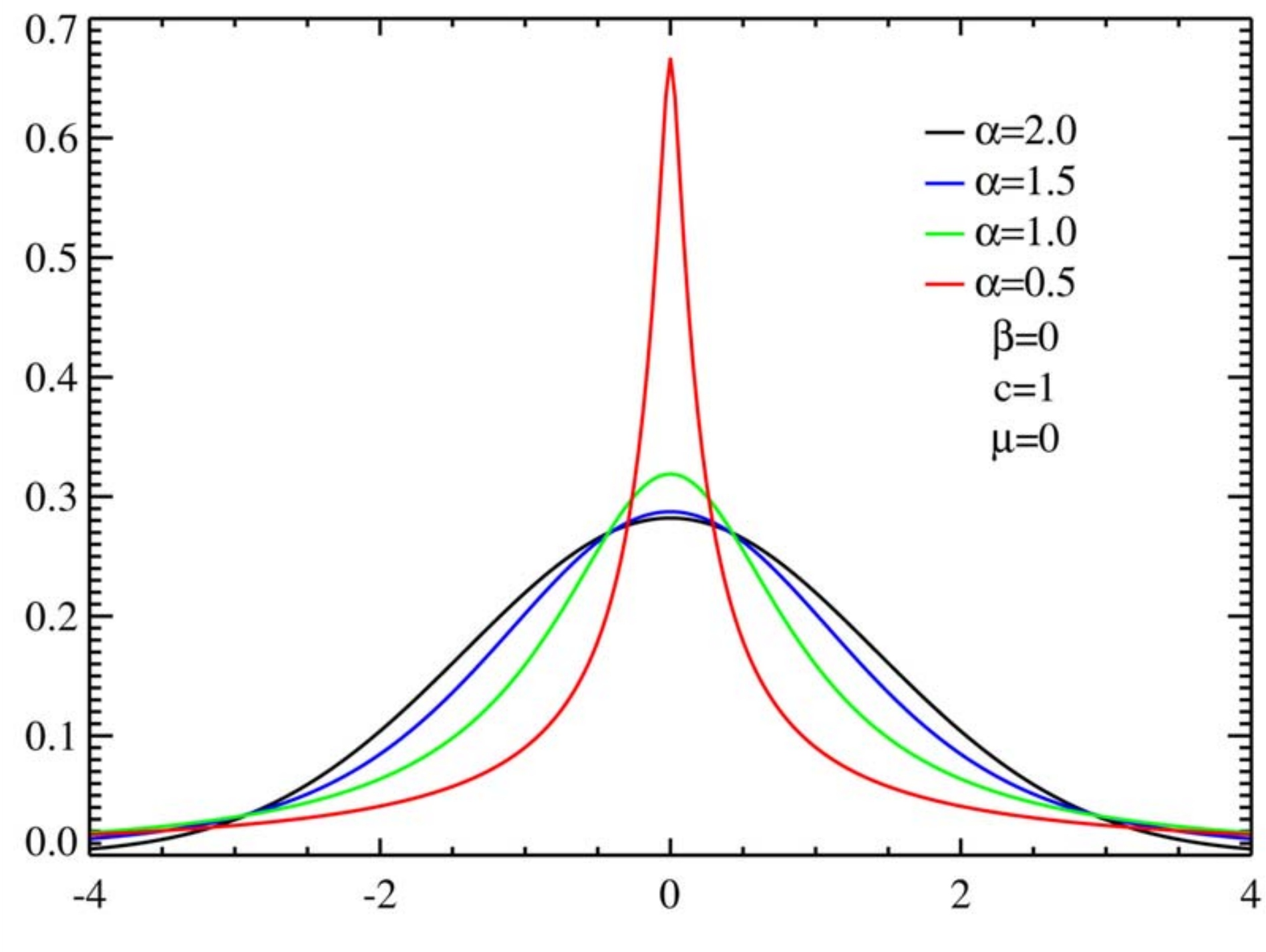}
\caption{\textbf Comparison of Levy Probability Distribution for different value of $\alpha$
\label{fig:Levy_distribution}}
\end{figure}

P. Levy in 1930s discovered a class of probability distribution which has an infinite second moment and governing the sum of these random variables [1]. The process is called stable, if the sum of these variables has the same probability distribution as individual random variables. A typical example of the stable process is the Gaussian process. While the Gaussian process has a finite second moment. The stable probability distribution that has an infinite second moment is then called the Levy probability distribution and has the following form [1]:
\begin{equation}
P_\alpha,_\epsilon(l)=\frac{1}{\pi}\int_0^{10} e^{{-\gamma{q}}^{\alpha}} cos({q}{l}) dq
\end{equation}
The distribution is symmetry with respect to $l=0$, $\gamma$ is the scaling factor and $\alpha$ determines the shape of the distribution. The required value of $\alpha$ is between 0 and 2. The parameter $\alpha$ determines the shape of the distribution in such a way that different shapes of probability distribution in the tail region can be obtained. The bigger is the parameter $\alpha$, the shorter the tail region. In the limit of alpha is $2$, the distribution will become the Gaussian distribution and no longer Levy distribution. Thus, by fixing $\gamma=1$, for large values of $l$, (1) can be approximated by [1],[4]:
\begin{equation}
P_\alpha(l) {\approx}{ } l^{-\alpha}
\end{equation}
Viswanathan et al. [4] derived several equations in order to optimize the Levy flight's parameters. They describe that the mean number $N$ of flights (movements) between successive target sites is approximated by:
\begin{equation}
N \approx (\frac{\lambda}{r_v})^{\frac{(\alpha-1)}{2}},
\end{equation}
where $\lambda$ is the average distance between two successive target sites and $r_v$ is the sensing range of the forager. They also describe that the optimal efficient value of alpha can be approximated by:
\begin{equation}
\alpha = 2 - \beta,
\end{equation}
where
\begin{equation}
\beta \approx  \frac{1}{(ln \frac {\lambda}{r_v})^2}.
\end{equation}
So in the absence of a priori knowledge about the distribution of target sites, an optimal strategy for a forager is to choose $\alpha=2$ when $\frac {\lambda}{r_v}$ is large but not exactly known.

\section{Artificial Potential Field Method}
\label{sec:petentF}

Artificial potential field creates a field, or gradient, among robots and their environment. This method was originally invented and introduced by Khatib [17] for robot manipulator path planning and is widely used in many variants in the robotics. The basic idea of the potential field approaches is that the robot is attracted towards the goal, i.e. searching target, or object of interest, while being repulsed by the obstacles that are known in the environment. The superposition of all forces is applied to the robot and smoothly guides the robot toward the goal while simultaneously avoiding known obstacles. In this paper, we proposed the generating repulsion forces among robots to improve the dispersion process during deployment.

If we assume a differentiable potential field function $U(q)$, we can find the related force $\underline{F}(q)$ at position $q=(x,y)$
\begin{equation}
\underline{F}(q)= -\underline{\nabla}{U(q)},	
\end{equation}
where $\underline{\nabla}{U(q)}$ denotes the gradient vector of $U$ at position $q$.
\begin{equation}
\partial{U}= \begin{bmatrix}
{\frac{\partial U}{dx}}\\
							\\
{\frac{\partial U}{dy}}\\
\end{bmatrix}
\end{equation}
The idea of the repulsive potential is to generate a force among robots to repel each other. This repulsive potential should be very strong when the robot is close each other, but should not influence their movement when they are far away from each other. One example of such a repulsive field is [17]:
\begin{equation}
{U}_{rep}(q)=\left\{ \begin{array}{rl}
\frac{1}{2} k_{rep}({\frac{1}{\rho{(q)}}} - {\frac{1}{\rho_{0}}}) & \text{if } \rho(q) \geq \rho_0,\\
0 & \text{if } \rho(q) < \rho_0,
\end{array} \right.
\end{equation}
where $k_{rep}$ is a scaling factor, $\rho(q)$ is the minimal distance from q to the adjacent robot and $\rho_0$ the threshold value of the distance. The repulsive potential function $U_{rep}$ is positive of zero and tends to infinity as $q$ gets closer to the other robot. This leads to the repulsive force:
\begin{equation}
\label{eq:rfe}
\underline{F}_{rep} = - \underline{\nabla} U_{rep}(q)=
\end{equation}
\begin{equation*}
        = \left\{ \begin{array}{rl}
k_{rep}({\frac{1}{\rho{(q)}}} - {\frac{1}{\rho_{0}}}) \frac{1}{\rho^2(q)} \frac{\underline{q}-\underline{q_{neighbor}}}{\rho(q)} & \text{if } \rho(q) \geq \rho_0,\\
0 & \text{if } \rho(q) < \rho_0.
\end{array} \right.
\end{equation*}
Thus, the amount of the repulsive force will accelerate the robot movement to the opposite direction of the potential field source.

From the repulsion force equation (\ref{eq:rfe}), it is also concluded that the robot is required to have an active sensing capability to measure its distance with the other robots.

\section{Implementation Framework}
\label{sec:imp}

\subsection{Generic Framework of the Implementation}

One main goal of this work is to implement a generic implementation framework of the Levy flight random walk in the robotic platform. The generic implementation framework is expected to be able to simplify the modification of the search algorithm and parameters during the experiment. Performance investigation can be performed by comparing the result of the implementation with some usual search methods.  Nevertheless, the framework should also simplify the calibration process in the real robot implementation. Therefore, simple kinematic equation to model the robot movement is derived:

\begin {equation}
	\begin{bmatrix}
	\dot{x(t)} \\
	\dot{y(t)} \\
	\dot{\theta(t)} \\
	\end{bmatrix}
	=
	L(t) \begin{bmatrix}
	k cos \theta(t) \\
	k sin \theta(t) \\
	0
	\end{bmatrix}
	+ (1-L(t)) \begin{bmatrix}
	0 \\
	0 \\
	\omega(t)
	\end{bmatrix}
\end {equation}

\begin{figure}[ht]
\subfigure[a]{\includegraphics[width=.22\textwidth]{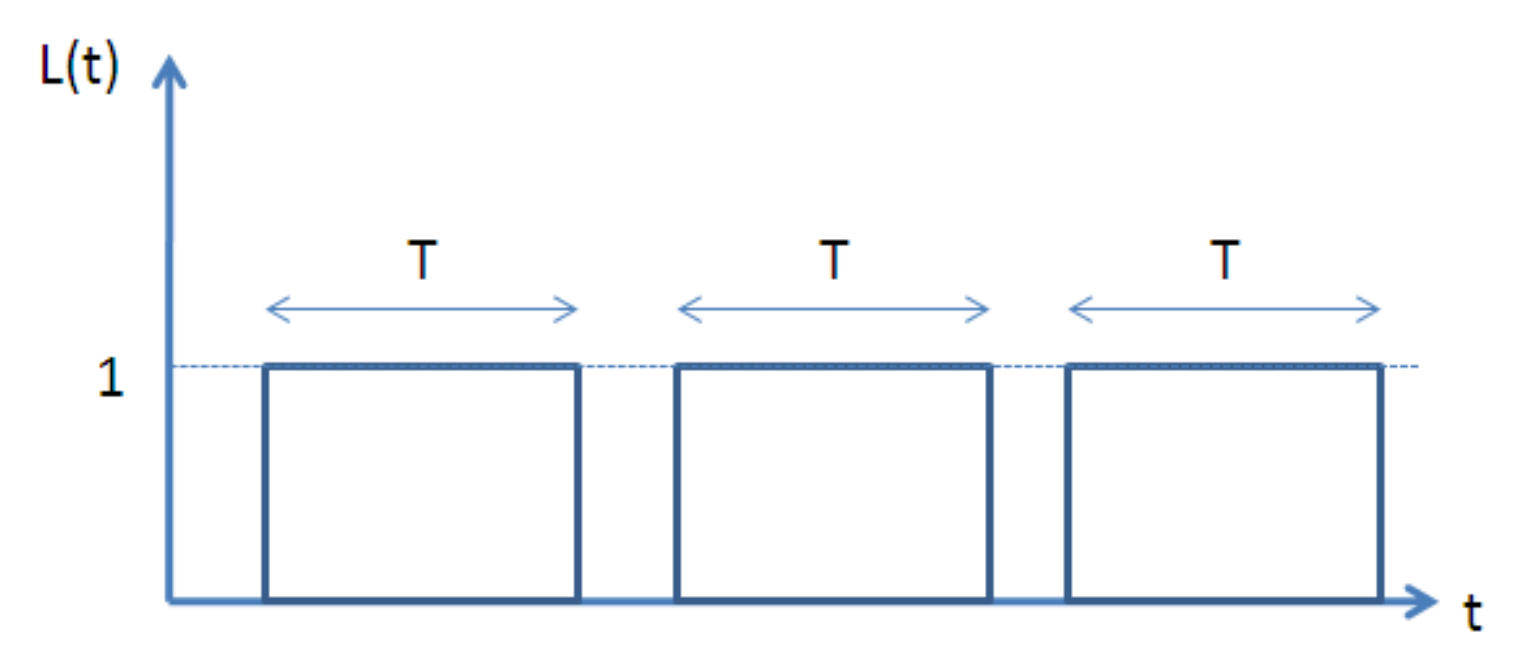}}~
\subfigure[b]{\includegraphics[width=.22\textwidth]{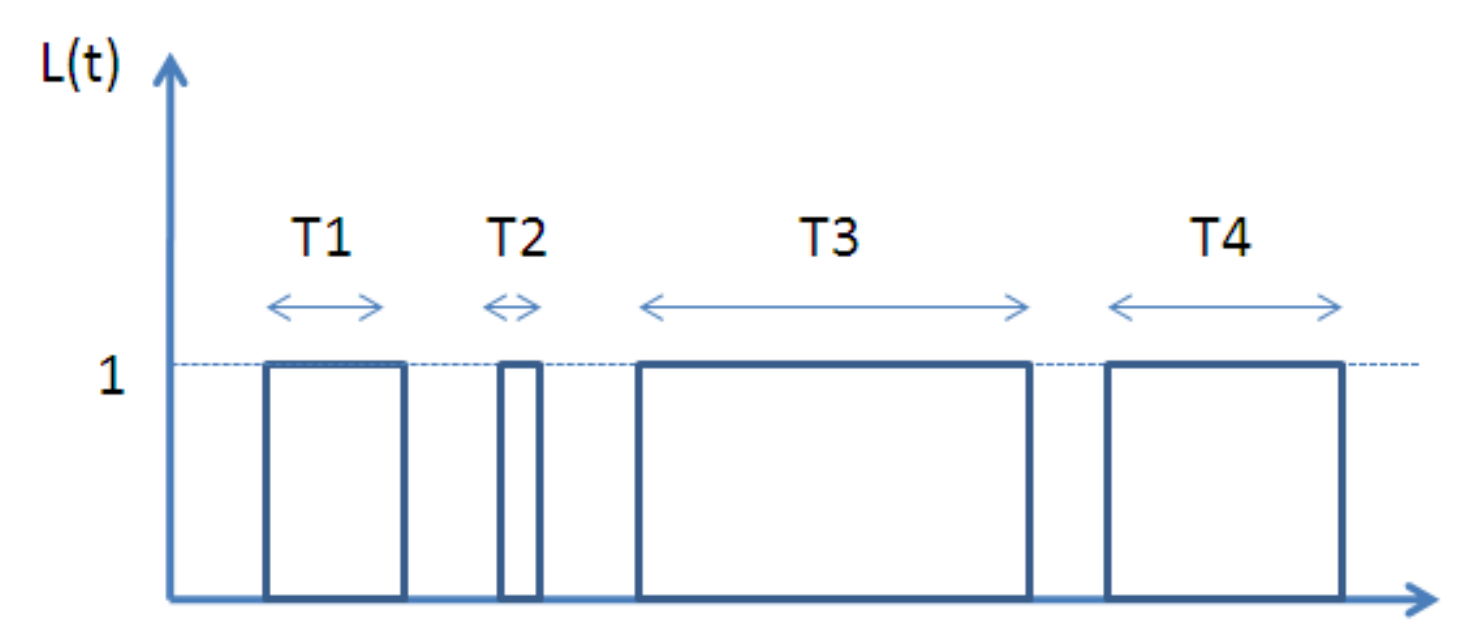}}
\caption{\textbf{(a)} Waveform for fixed-length random walk.
\textbf{(b)}Waveform for Levy-flight random walk.
\label{fig:oscillator}}
\end{figure}

Here, $x(t)$ and $y(t)$ are position of the robot at time $t$ in Cartesian coordinate, $\theta(t)$ is the orientation of the robot, $k$ is the constant linear velocity, $L(t)$ is the control value to turn on and to turn off the movement of the robot, and $\omega(t)$ is the angular velocity of the robot randomized by the normal Gaussian distribution function. Within this equation, the period of $L(t)$ performs the length of the robot walk and the value $\omega(t)$ determines the orientation movement of the robot. The value of $L(t)$ comes from the square wave oscillator output, where the waveform period is determined by the type of the random walk. The period of $L(t)$ and $\omega(t)$ value are parameters to be modified during experiments. This principle is bio-inspired controller implementation similar to Central Pattern Generator (CPG) that can be found in some animals locomotion mechanisms and inspired by [2] for bacterial chemotaxis foraging implementation.
\begin{figure}[ht]
\centering
\subfigure{\includegraphics[width=.42\textwidth]{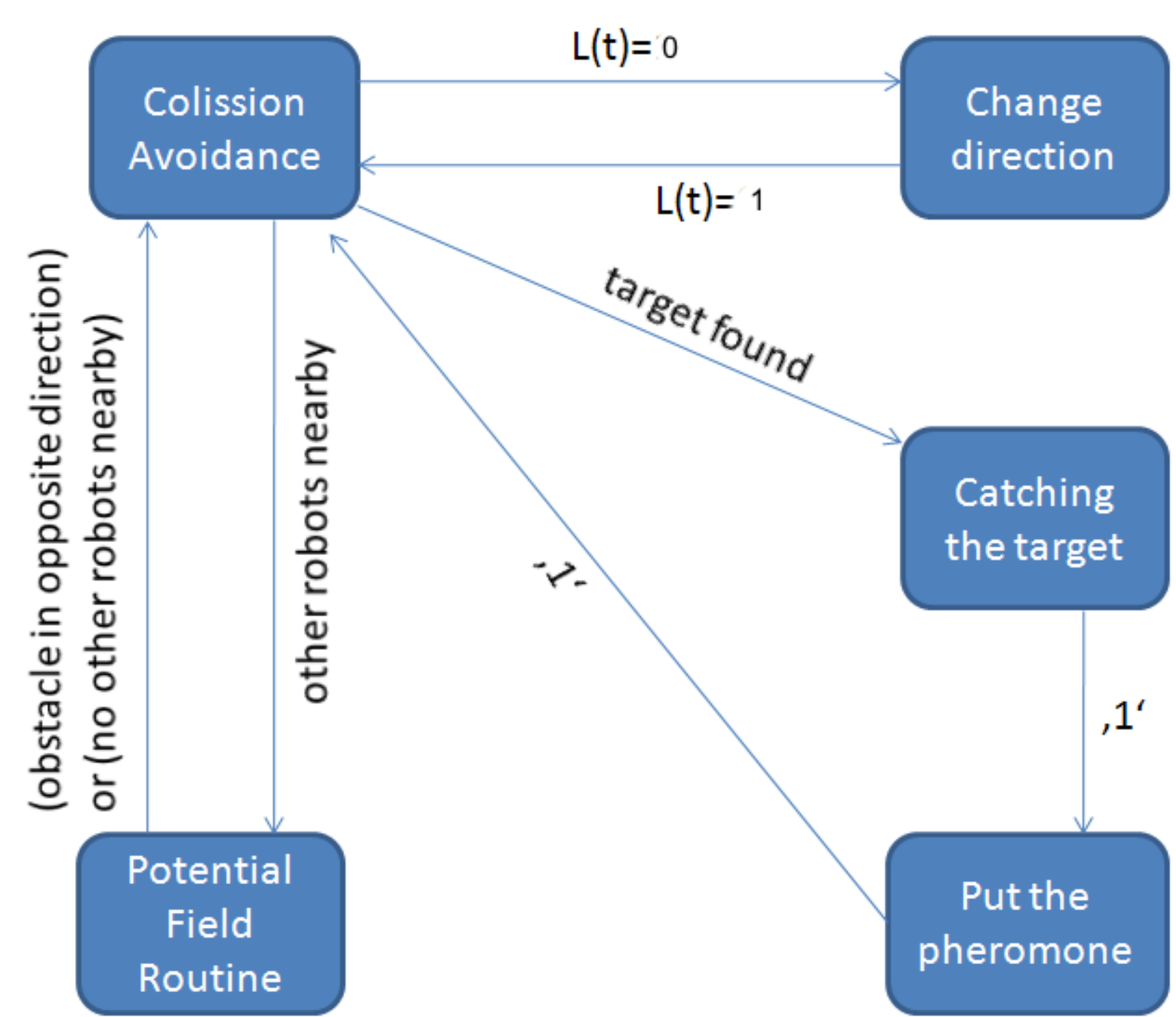}}
\caption{Finite State Machine Model of the Controller.\label{fig:fig4}}
\end{figure}

As described before, the square wave oscillator output drives the activation of the locomotion system of the robot. Since the possible value of the driver signal is only '1' and '0', a simple 'on-off' controller is required. Fig.~\ref{fig:fig4} shows the Finite State Machine (FSM) model of the designed controller. During the initial state of the controller, robot performs the collision avoidance action. After the first positive period, the value of the $L(t)$ becomes '0'. The robot changes the direction during the 'zero' phase within $\omega(t)$ angular velocity. The length of the positive phase is defined by the result of the Levy probability function generator or the period of the constant frequency oscillator. The robot will enter the 'Collision Avoidance' state again as soon as the value of $L(t)$ becomes '1'. If there is another robot in its sensing range, the controller state will move to the 'Potential Field Routine'. In this routine, controller will measure the distance to the adjacent robot for calculating the repulsion force, and then accelerate the movement of the robot to the opposite direction of the adjacent robot within the calculated repulsion force. Finally, if the robot found the object of interest during the robot the 'Collision Avoidance' state , it will enter the 'Catching Target' and will put the artificial landmark with repelling 'pheromone'. Here, the pheromone is implemented as a specific message content to the target to turn on the 'repelling signal'. Therefore, other robots, which are aware of the 'repelling signal', will avoid the area around the found target and will disperse to other foraging area.

\subsection{Levy Flight Implementation}

An algorithm for generating random numbers based on Levy probability distribution is needed to determine the length of the walk of the robot during the foraging phase. Such algorithm was introduced in [1]. This algorithm requires two independent random variables $a$ and $b$ which have a normal Gaussian distribution from this nonlinear transformation
\begin{equation}
v=\frac{a}{{\vert b \vert}^{\frac{1}{\alpha}}}
\end{equation}
within the nonlinear transformation, the sum of variables with an appropriate normalization
\begin{equation}
z_n=\frac{1}{n^{\frac{1}{\alpha}}}\sum_{k=1}^n{v_k}
\end{equation}
converges to the Levy probability distribution with larger $n$ (the usual value of $n$ is 100 [1][6]).

Since Gaussian random number is required for the Levy flight implementation and for randomizing the direction of moving, an appropriate Gaussian approximation is investigated. Box-Muller transformation [8] is chosen in the implementation, since it is a well known numerical approximation for generating Gaussian distribution random number and has many available programming implementation. Box-Muller transformation is popular, because it is simple and fast for high level language implementation [9].

\subsection{Robotic Simulation Platform}

Simulation platform is required, because the real robotic platforms do not have a global localization system and also have complex kinematic models. Simulation platform can simplify the modification of control parameters and environmental condition. Additionally, it is also more scalable, thus makes some experiments with hundreds of robots possible. Webots from Cyberobotic is chosen as the simulation platform.

\begin{figure}[ht]
\centering
\subfigure[]{\includegraphics[width=.2\textwidth]{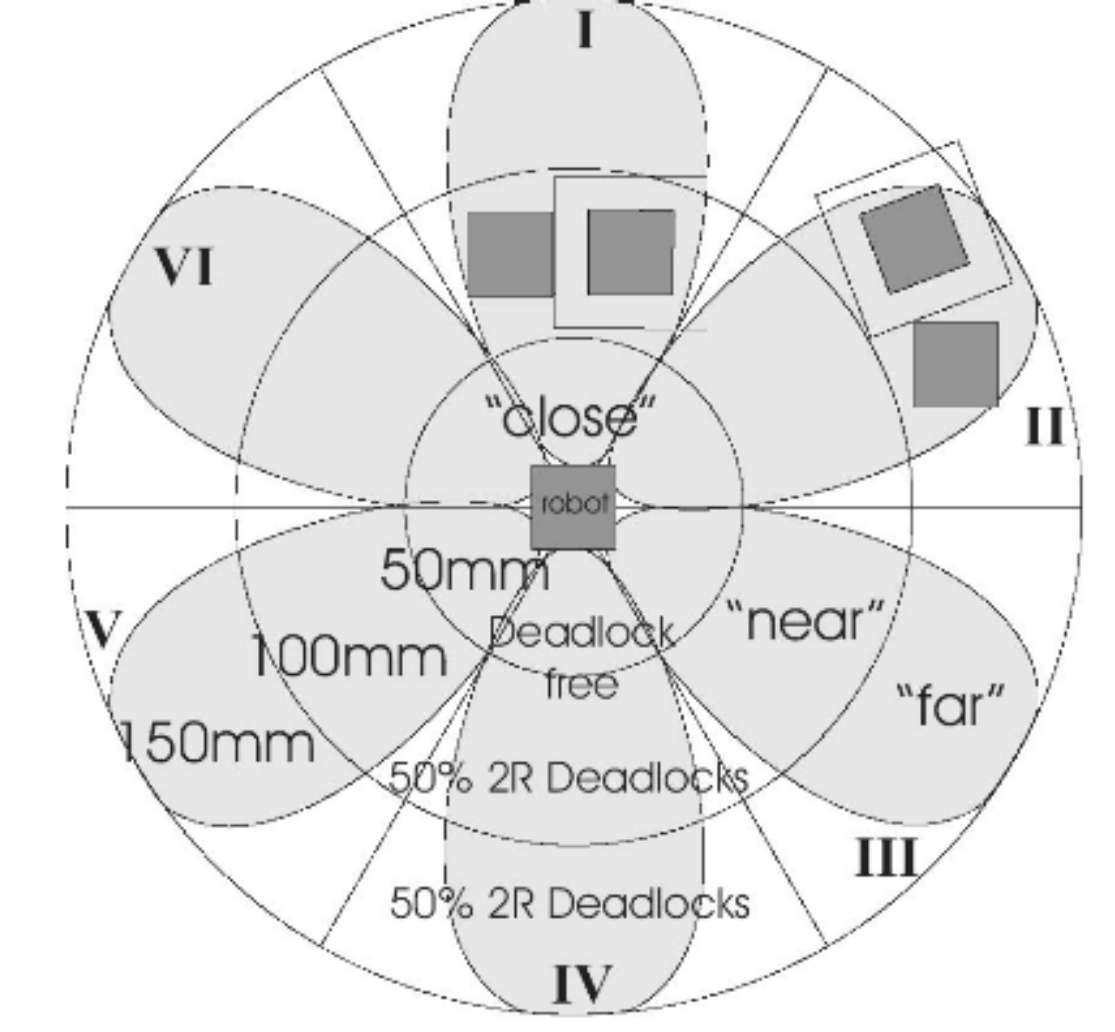}}~
\subfigure[]{\includegraphics[width=.28\textwidth]{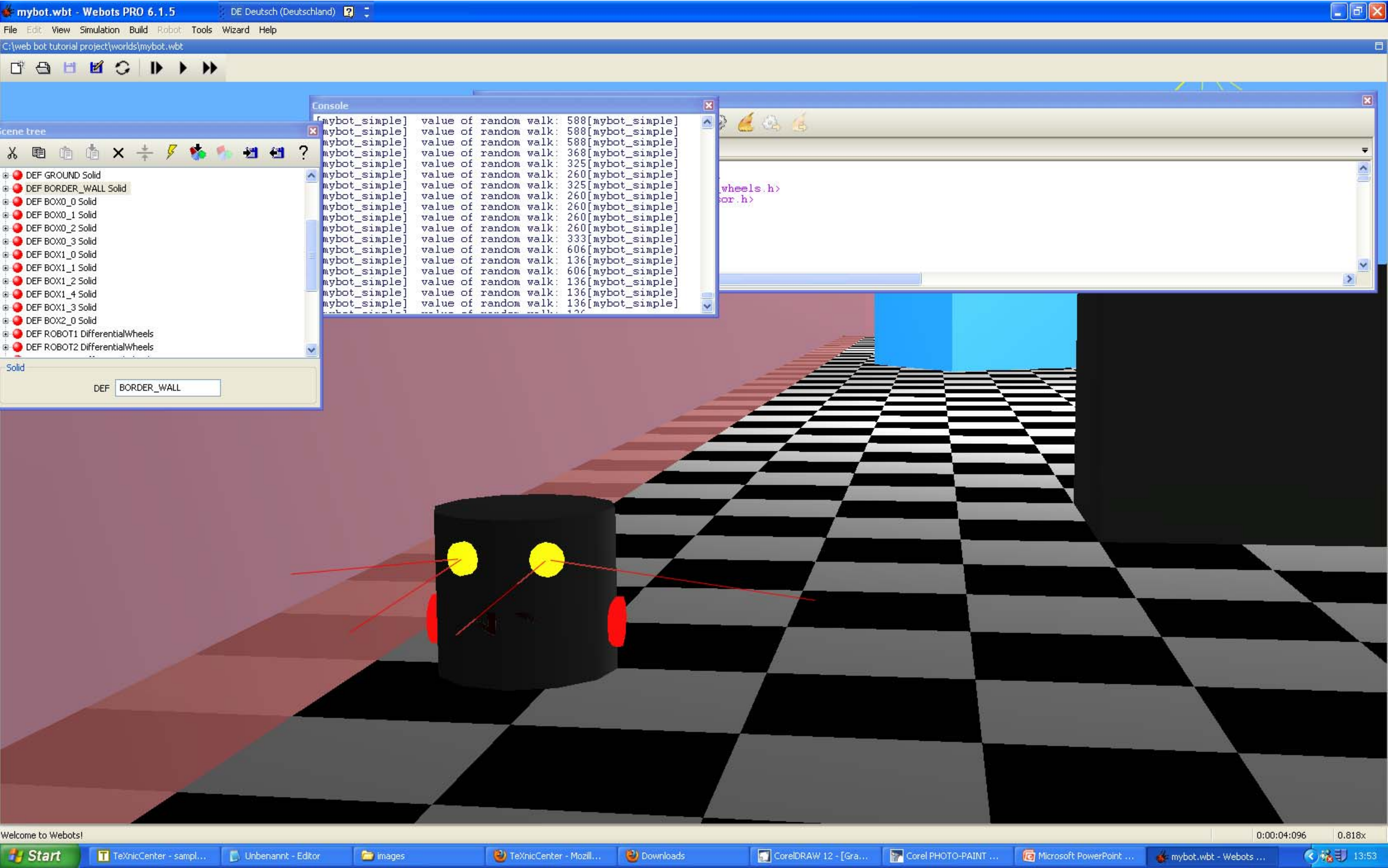}}
\caption{\textbf{(a)} Radiation Pattern of Sensing for Jasmine [10].
\textbf{(b)} Robotic platform in the simulation.
\label{fig:simulation_robot}}
\end{figure}

For surface application, later the random search algorithm will be implemented in the Jasmine robot. Therefore, the robotic platform in the simulation must have similar kinematic model and sensing capability with the real Jasmine, see Fig.~\ref{fig:simulation_robot}. Jasmine uses six IR sensor for active sensing and communication. In the locomotion part, two differential wheels with two DC motors are located on the right and on the left side of the Jasmine robot [21]. Therefore, similar sensor and locomotion mechanism are also realized in the simulation robot.

\section{Multi-Robotic Random Search Experiments}
\label{sec:exp}

\subsection{Experimental Setup}

\begin{figure}[ht]
\centering
\subfigure{\includegraphics[width=.4\textwidth]{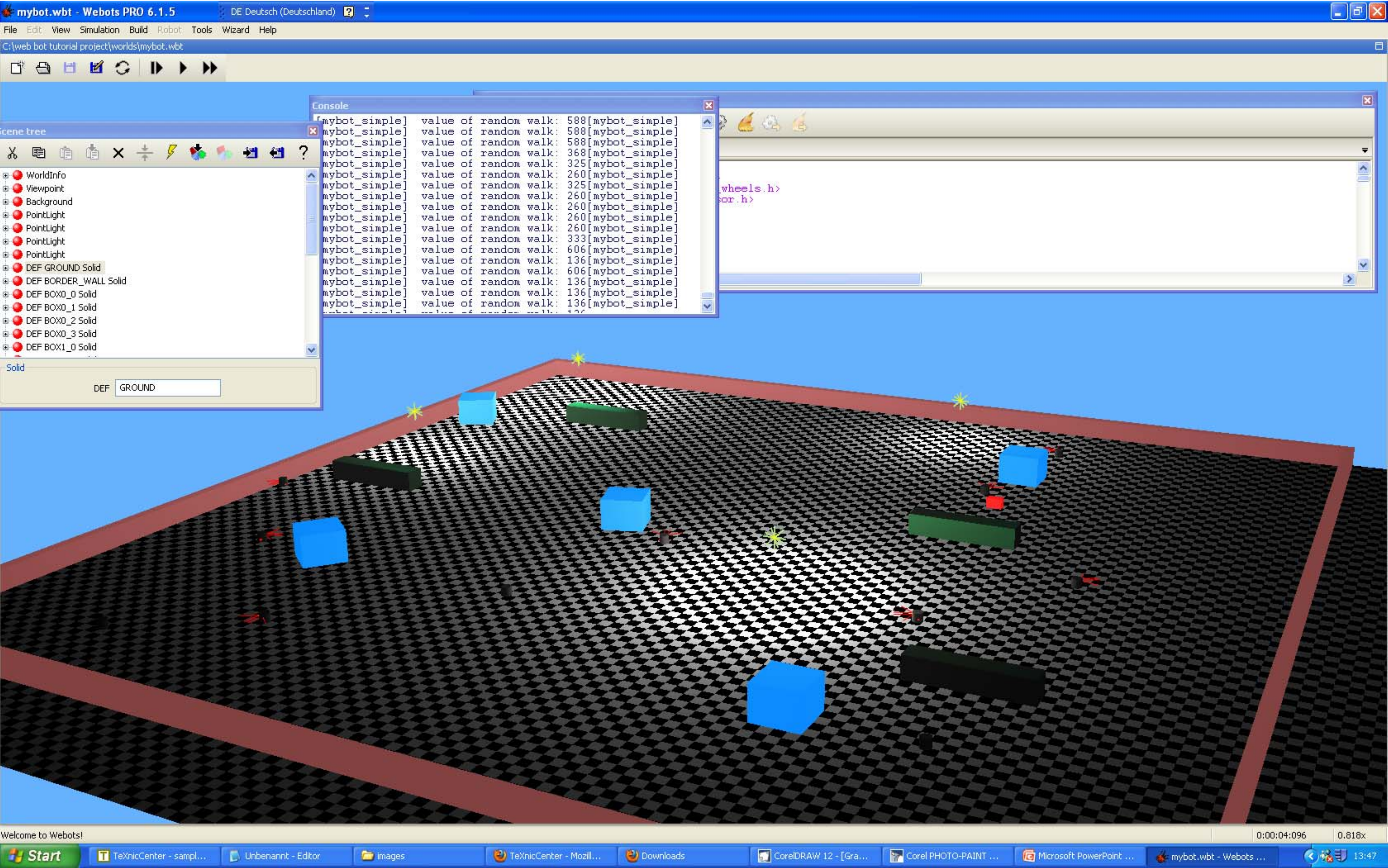}}
\caption{\textbf Simulation Arena.
\label{fig:arena}}
\end{figure}

Big simulation arena that emulates the environment is necessary to implement swarm robotic experiments with many robots. Therefore, an arena that has 20$\times$20 simulative 'meters' of size is prepared. Several obstacles with different size and shape are placed randomly on the simulation arena, see Fig.~\ref{fig:arena}. Nevertheless, objects of interests as searching targets are also prepared. The object of interests is a passive static robot that transmit a specific message within the same range as robot sensor.

\begin{figure}[ht]
\centering
\subfigure[]{\includegraphics[width=.24\textwidth]{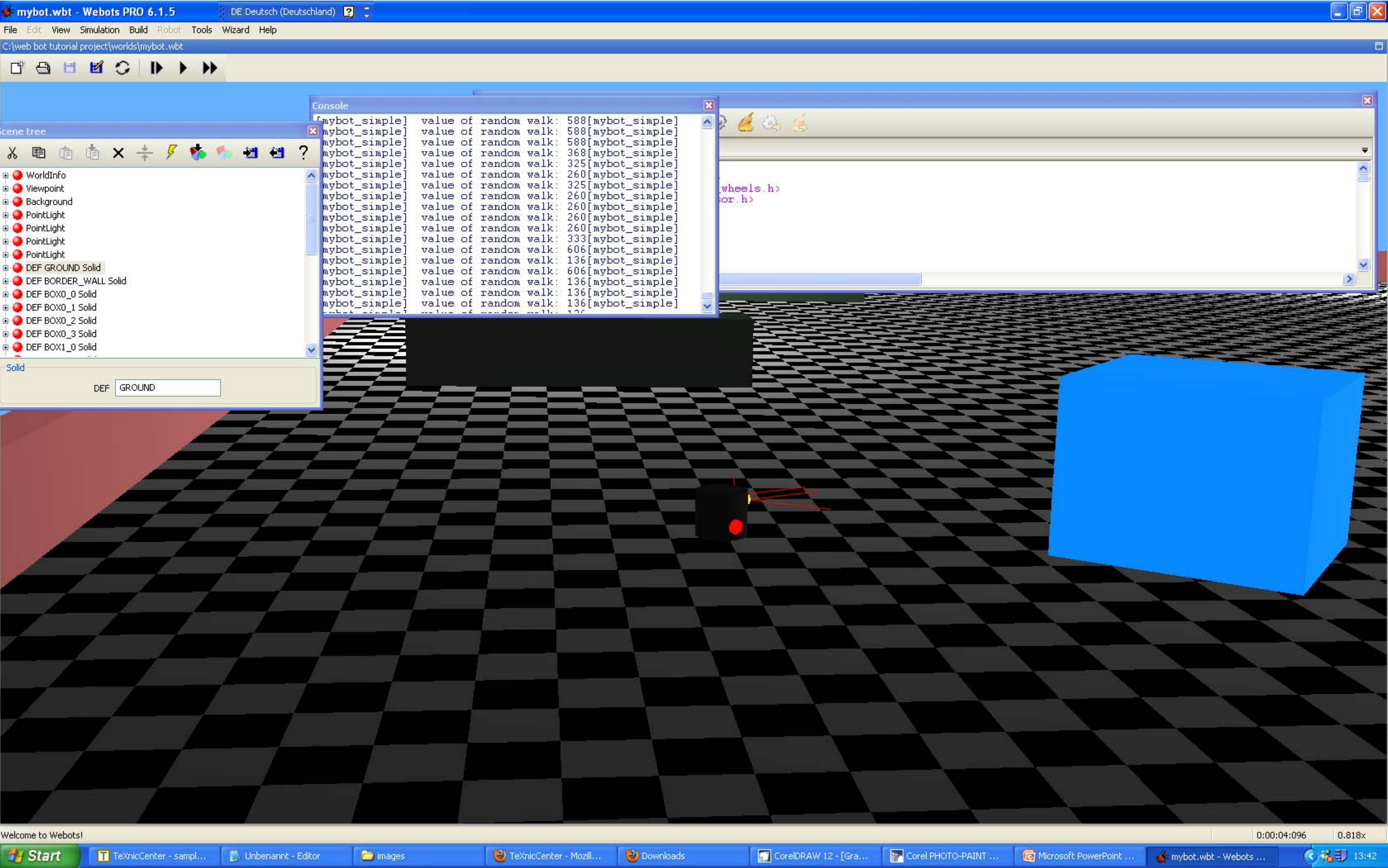}}~
\subfigure[]{\includegraphics[width=.24\textwidth]{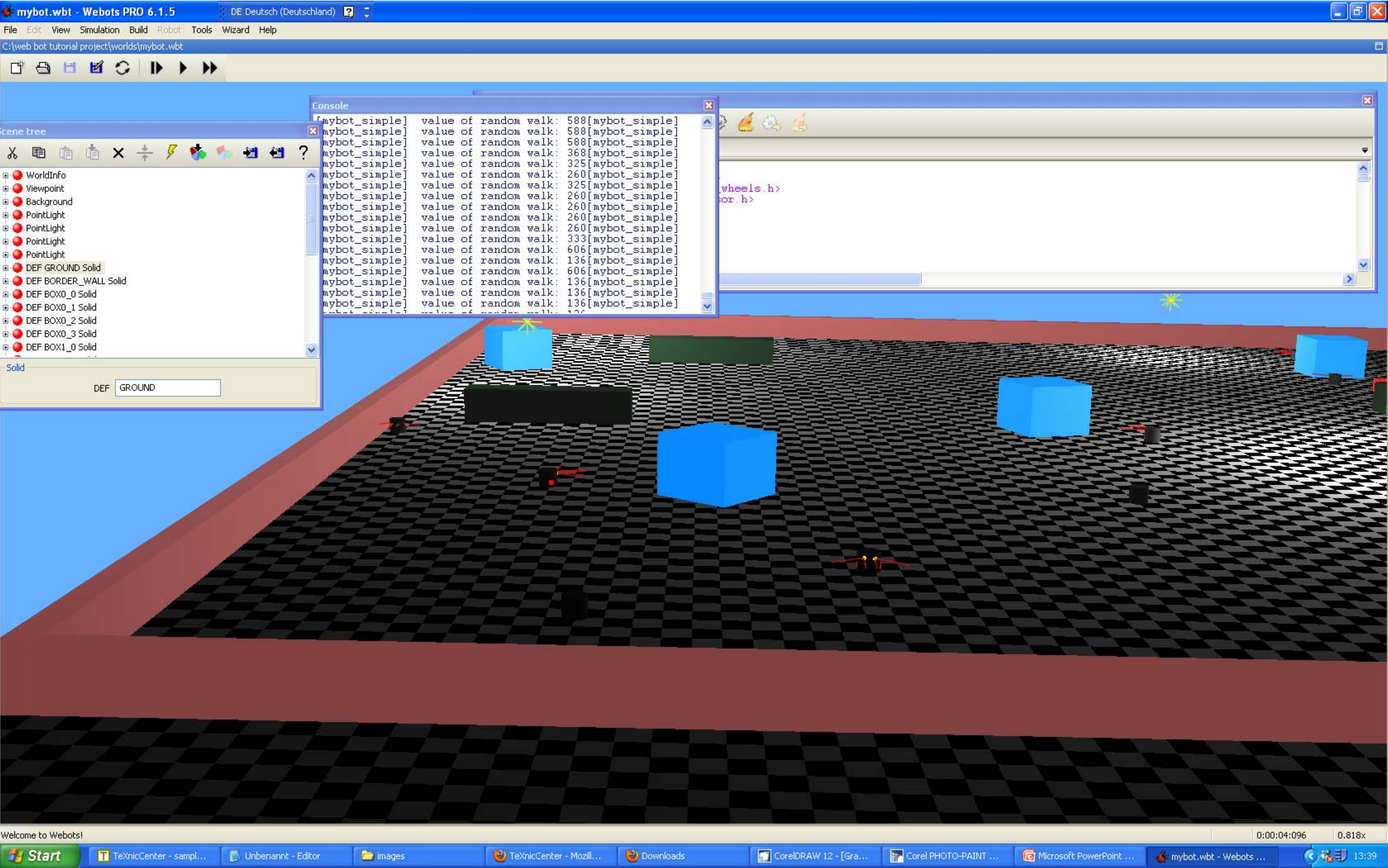}}
\caption{\textbf{(a)} Single robot random search in the simulation.
\textbf{(b)} Multi robot random search in the simulation.
\label{fig:simulation_robot2}}
\end{figure}

The main purpose of the experiments is to measure the minimum time travel that is required by all robots to find all of the foraging targets. Several experiments performed in this paper include comparison for different numbers of robots in the searching mission, comparison with other random walk algorithm, and investigation related with the application of the artificial potential field, see Fig.~\ref{fig:simulation_robot2}. The other random walk algorithm that is compared with the Levy flight in our experiment is fixed-length random walk.
	
From the generic framework, it is described that several variable values can be modified during the experiments and others remained constant. The constant linear velocity $k$ is configured as 60 cm/sec. The period of oscillator value to control the transition of the walk and to stop the phase is taken from the output value of the Levy distribution generator. It can be also replaced by a constant frequency oscillator for the fixed-length walk mode. Therefore, the period of the oscillator determines the length of the robot walk in the recent phase, and the frequency of changing movement direction.

\subsection{Experimental Result}

During the experiments, several individual robots are randomly deployed in the arena to search the targets. For each experiment, different number of robots and targets are performed. Every experiment is executed in 20 trials on different environmental conditions (e.g. position of the targets and position of the obstacles).

Since there are tens of identical robots deployed together in the environment, different swarm-like strategies can be explored.  Intuitively, by using more robots in the searching process, the searching time will reduce significantly due to the increasing number of robot. However, the collision avoidance mechanism allows the robot to avoid each other and this affects the pattern of the length of walk. Every time the robots meet each other, robot will change its direction earlier then expected, this reduces the required length of walk determined by the Levy distribution function. Therefore, reducing searching time by increasing the number of robots will be saturated in some points.

In order to perform collective behavior, robot must have communication peripheral to propagate information regarding the individual progress to other robots. Swarm communication can be performed in many different ways, i.e., by using direct communication, by using the environment (stigmergy), or by using bio-inspired pheromone [19]. In this experiment, pheromone based approach is applied for information sharing among robot. If one robot can find a target, it will leave a signature on the target, so that the other robot will aware to this signature and disperse to the rest area of the environment to find other targets. The awareness of the robot to recognize the active pheromone is wider than the sensing range of the robot. Nevertheless, instead of only leaving signature to the found target, other useful information might also be shared. Furthermore, it is nice to have a signature on the found target, since it simplifies the observation during the experiments.
\begin{figure}[ht]
\centering
\subfigure{\includegraphics[width=.48\textwidth]{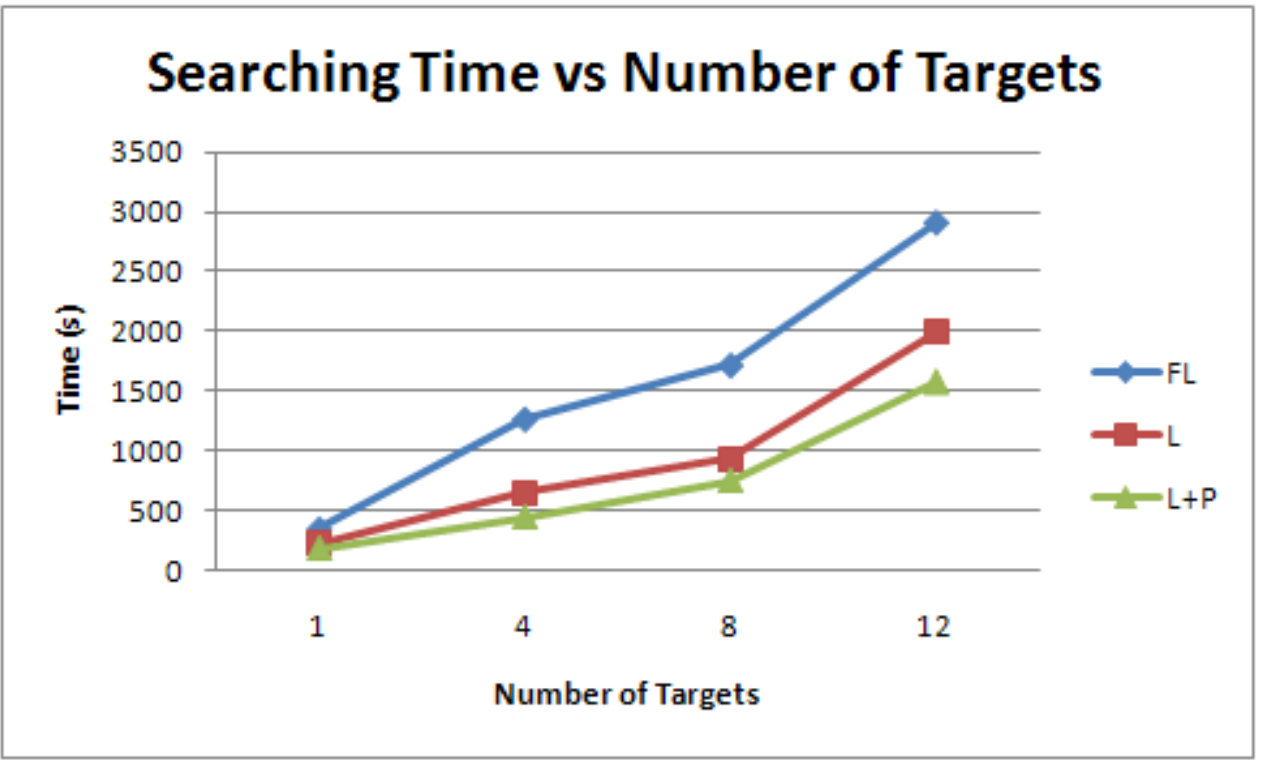}}
\caption{\textbf Multi robot random-searching experiments, searching time vs number of targets, number of robots: 10, L+P: Levy random-walk and Potential Field, L: Levy random-walk, FL: Fixed-length random}
\label{fig:chart2}
\end{figure}

\begin{figure}[ht]
\centering
\subfigure{\includegraphics[width=.48\textwidth]{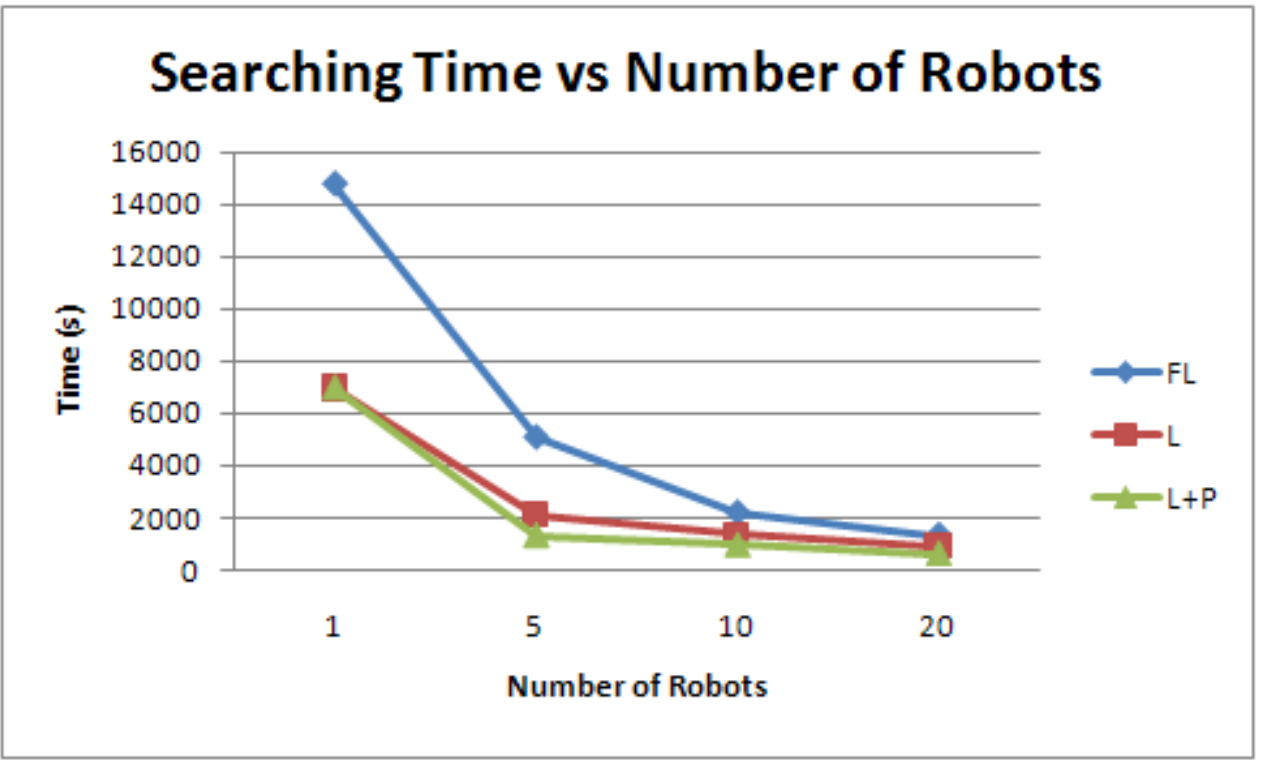}}
\caption{\textbf Multi robot random-searching experiments, searching time vs number of robots, number of targets:8, L+P: Levy random-walk and Potential Field, L: Levy random-walk, FL: Fixed-length random}
\label{fig:chart1}
\end{figure}

From the experiment results in Figs.~\ref{fig:chart2} and \ref{fig:chart1}, it can be seen that the Levy flight is effective for different numbers of robots. However, if the number of robots exceed some limit, its effectiveness has essential different with the fixed-length movement. The large number of robots forces an individual robot to be more frequent in changing direction to avoid other robot during foraging. Therefore, if changing a movement direction to avoid other robots becomes too frequent, method for determining the length of the walk has less impact on the searching performance. Nevertheless, comparing to the fixed-length random search, the Levy flight is still more effective for multi-robot application in every experimental condition. Finally, it can be shown that the implementation of the artificial potential field among robots also increase the performance of the searching algorithm.

\section{Conclusion and Future Work}
\label{sec:conlusion}

In this paper, we investigated an efficient random search algorithm for multi-robot application that is crucial in exploration of large areas. The proposed idea consists in using the Levy flight algorithm for determining the length of the walk and artificial potential field to improve the efficiency of the dispersion during deployment. The approach has the advantage that it does not require centralized control or localization system, and will therefore scale the possibility to apply very large number of robots. Targeted scenario is primarily related to underwater exploration, however the search and rescue scenario with swarm-like robots (e.g. aerial surveillance) can also be envisaged. It is also proposed to build a generic implementation framework to be effectively executed both for simulation and for real robot applications.

Experiment results showed that the Levy flight algorithm can achieve a better performance compared to the fixed-length random walk. Experiment results also demonstrated that the algorithm becomes more efficient by applying the artificial potential field. However, if the number of robots exceed a limit, the effectiveness of the algorithm has less impact to the whole performance. Several open questions are related to a physical implementation of landmarks with artificial pheromone and their dropping/collecting in underwater environment.

Further works are primarily related to implementation on surface with Jasmine and further transition to the underwater platform. After finding and verifying an appropriate algorithm for a random search, the framework, algorithm and parameters must be ported to the real robot implementation. The Jasmine robots are envisaged because of real 2D environment and a possibility to explore the impact of real constraints on random-search approach. Later it will be implemented on the ANGELS test platform for underwater foraging and ecological applications.

\section*{Acknowledgement}

This work is supported by FP7 EU Project  ANGELS - ANGuliform robot with ELectric Sense, the grant agreement no. 231845.  Additionally, we want to thank all members of the projects for fruitful discussions.

\end{document}